\pdfoutput=1

\documentclass[11pt]{article}

\usepackage[]{acl}

\usepackage{times}
\usepackage{latexsym}
\usepackage{multirow}

\usepackage[T1]{fontenc}

\usepackage[utf8]{inputenc}

\usepackage{microtype}

\usepackage{inconsolata}

\usepackage{graphicx}

\usepackage{hyperref}       
\usepackage{url}            
\usepackage{booktabs}       
\usepackage{amsfonts}       
\usepackage{nicefrac}       
\usepackage{microtype}      
\usepackage{xcolor}         
\usepackage{xspace}
\usepackage{booktabs} 

\newcommand{\llama}{\textsc{\textbf{llama}}\xspace}
\newcommand{\gpt}{\textsc{\textbf{gpt}}\xspace}

\title{Open or Closed LLM for Lesser-Resourced Languages?\\ Lessons from Greek}

\author{John Pavlopoulos \\
  AUEB \& Archimedes/Athena RC, Greece \\
  \texttt{annis@aueb.gr} \\\And
  Juli Bakagianni \\
  University of Ioannina, Greece\\\AND
  Kanella Pouli\\ILSP/Athena RC, Greece\\\And
  Maria Gavriilidou\\ILSP/Athena RC, Greece
\\}

\begin{document}
\maketitle
\begin{abstract}Natural Language Processing (NLP) for lesser-resourced languages faces persistent challenges, including limited datasets, inherited biases from high-resource languages, and the need for domain-specific solutions. This study addresses these gaps for Modern Greek through three key contributions. First, we evaluate the performance of open-source (Llama-70b) and closed-source (GPT-4o mini) large language models (LLMs) on seven core NLP tasks with dataset availability, revealing task-specific strengths, weaknesses, and parity in their performance. Second, we expand the scope of Greek NLP by reframing Authorship Attribution as a tool to assess potential data usage by LLMs in pre-training, with high 0-shot accuracy suggesting ethical implications for data provenance. Third, we showcase a legal NLP case study, where a Summarize, Translate, and Embed (STE) methodology outperforms the traditional TF-IDF approach for clustering \emph{long} legal texts. 
Together, these contributions provide a roadmap to advance NLP in lesser-resourced languages, bridging gaps in model evaluation, task innovation, and real-world impact.
\end{abstract}

\section{Introduction}
\label{sec:intro}

Natural Language Processing (NLP) tasks have advanced significantly with the help of deep learning, and more recently with large language models (LLMs), the creation of which demands immense volumes of digital data \cite{brown2020language}. While multilingual NLP has benefited from these advances, the progress in lesser-resourced languages significantly lags behind that of well-supported languages. As a result, NLP for the myriad of languages worldwide relies heavily on research conducted for well-established languages, often inheriting their assumptions, biases, and other characteristics that may not align with the features of less supported languages \cite{bakagianni2024towards}. 

NLP for lesser-resourced languages is further hindered due to the scarcity of open-access, high-quality language resources. Such resources include datasets, which could be used to train and/or evaluate multilingual models on downstream tasks across languages or language models per se. Comprehensive NLP surveys for resource-lean languages, besides providing evidence on the digital readiness of the language, can also serve as a sound basis to promote NLP research for these languages, and to lead to available language resources for downstream tasks, opening the way to new benchmarks. Our work focuses on one of these languages, Greek,\footnote{In our study, with the term Greek we refer only to Modern Greek, excluding previous stages of the language's history.} the official language of Greece and one of the two official languages of Cyprus. 

In this work, based on the recent Greek NLP survey of \citet{bakagianni2024towards}, we extracted publicly available and open-access Greek datasets that permit derivatives, aggregating them into a unified collection that is easy to re-compile and use.\footnote{The code is available at: \url{https://github.com/greek-nlp/benchmark}}
Using this collection, we make three contributions:
\begin{itemize}
    \item  We benchmarked seven NLP tasks in Greek using a closed- (GPT-4o mini) and an open-source (Llama-70b) LLM, revealing task-specific strengths and weaknesses. \llama is better in Named Entity Recognition (NER) and Summarization while \gpt is better in Grammatical Error Correction (GEC), Machine Translation (MT; el-jpn), Intent Classification and Part-of-Speech (POS) Tagging. The two models perform on par on Toxicity Detection, MT (el-en, el-fa).  
    \item We reframe Authorship Attribution by evaluating LLMs in a 0-shot setting and hypothesising that high accuracy suggests potential inclusion of the authored texts in pre-training data. In lesser-resourced languages, which are not well supported in the training data, our hypothesis is even likelier.
    \item We introduce the first Text Clustering benchmark for Greek legal texts, demonstrating its value in organizing long, complex documents. Using the Summarize, Translate, and Embed (STE) methodology, we show improved clustering performance over a traditional baseline, offering a practical solution for navigating legal corpora in lesser-resourced settings.
\end{itemize}

\section{The data}\label{sec:method}
Based on the systematic literature review of \citet{bakagianni2024towards}, covering 142 research NLP studies for Greek from 2012 to 2023, we extracted and analysed information on the 94 available Greek datasets. Most of these resources were monolingual (84\%), but bilingual and multilingual datasets were also present (i.e., with Greek being one of the languages). The most frequent domains were politics and news (20\%) while the most frequent task was sentiment analysis (33\%). 

\begin{figure}[ht]
    \centering
    \includegraphics[width=.5\textwidth, trim=10 10 10 30, clip]{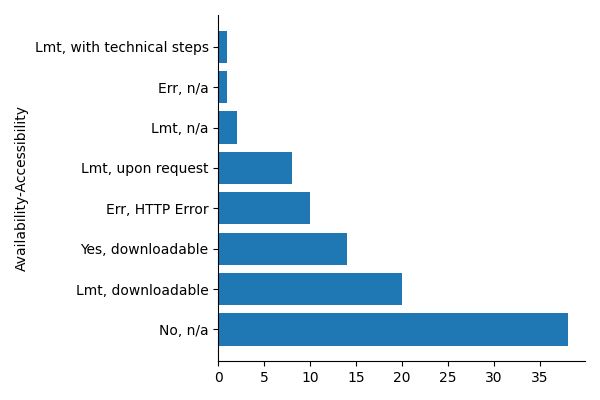}
    \caption{Availability and accessibility of extracted datasets, one tuple per bar. Availability is classified to open (yes), constrained (lmt), hindered by an error (err), or not available via a URL (no). Accessibility reflects the outcome we observed when accessing the resource.}
    \label{fig:availability}\vspace{-10pt}
\end{figure}

\subsection{Availability, Popularity and Licensing} 

\begin{figure}[ht]
    \centering\includegraphics[width=.5\textwidth, trim=60 0 60 30, clip]{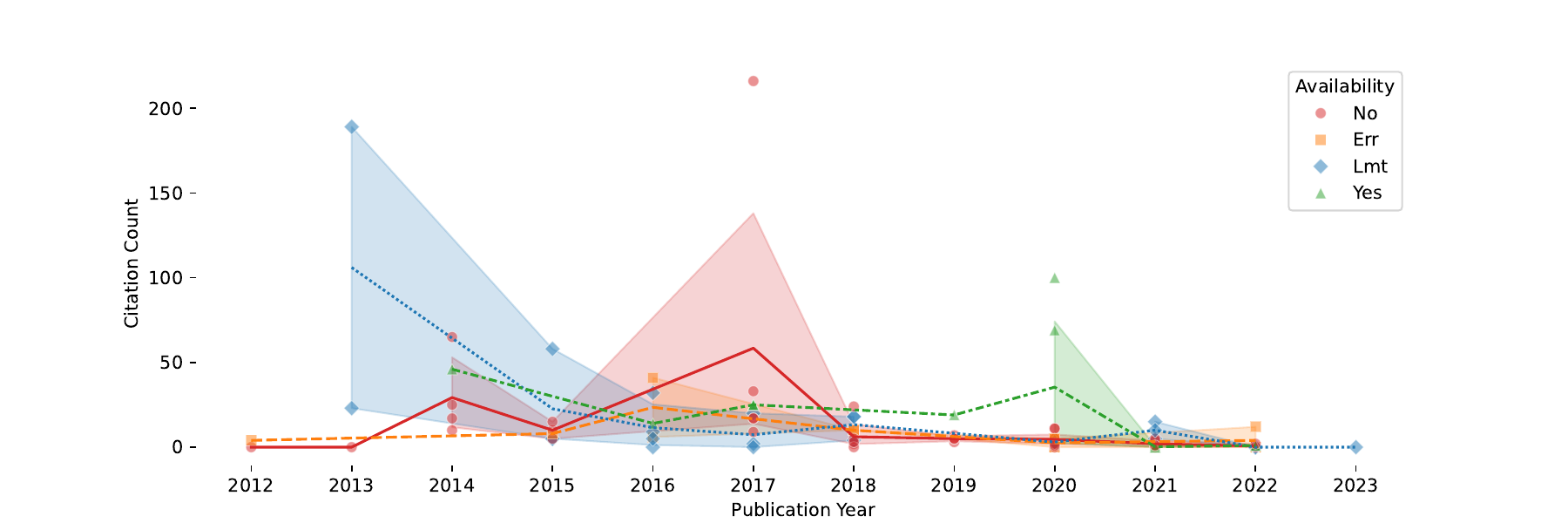}
    \caption{Citation counts per year of publication of studies developing datasets, based on the availability classified as yes, no, limited (lmt), or erroneous (err).}
    \label{fig:citation_ts}
    \vspace{-10pt}
\end{figure}

Figure~\ref{fig:availability} shows the availability status of the extracted datasets. The most common status is that the datasets are not available, meaning no information is provided regarding their availability. The second most frequent category concerns datasets reported as downloadable, but actually not publicly available due to restrictions, such as missing license or subscription limitations. 
\begin{figure}[ht]
    \centering
    \includegraphics[width=.5\textwidth,trim={5cm 0 3cm 0},clip]{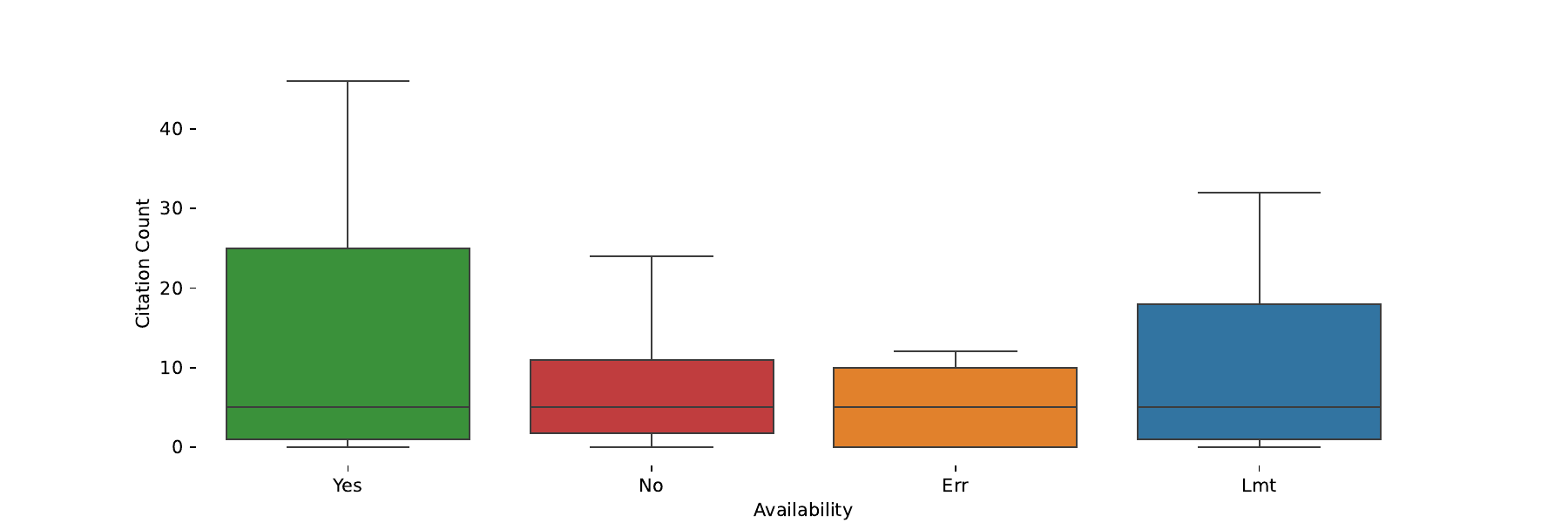}
    \caption{Whisker and box plot with the number of citations of studies per availability type.}
    \label{fig:citations}\vspace{-20pt}
\end{figure}
This limitation contradicts the principles and benefits of open access. Studies with publicly available datasets (serving as a reference point) can attract more citations, as is shown by the higher third and fourth quartile of the green box in Figure~\ref{fig:citations}. Historically (Figure~\ref{fig:citation_ts}), studies attracting citations while presenting data with limited \cite{juola2013overview}, or with no availability \citet{GIATSOGLOU2017}, belong to the past.
The study of \citet{juola2013overview}, for example, shown with a blue peak in 2013, concerns a shared task on authorship identification (hence, the citations) whose data lack a license (hence, the limitation). Our work reveals an alternative dataset for this task (\S\ref{ssec:authorship}). 

\subsection{Distilling FAIR data} 
We extracted findable, accessible, interoperable and re-usable (FAIR) datasets from the survey of \citet{bakagianni2024towards}. We focused on the 14 Greek annotated datasets that were found as licensed, accessible, machine-actionable, and with no hidden costs. We excluded the dataset of \citet{fitsilis2021development}, which is licensed under \href{https://creativecommons.org/licenses/by-nc-nd/4.0/deed.en}{CC BY-NC-ND 4.0} and does not allow derivatives (i.e., a limitation of re-usability). Also, \citet{korre2021elerrant} provided two datasets, one of which is too small in size (100 sentences). By disregarding these datasets, we end up with the 12 datasets presented in Table~\ref{tab:data}. There is a variety of domains and the largest dataset is that of \citet{Barzokas2020} followed by that of \citet{dritsa2022greek}.
\begin{table*}[ht]
  \centering\resizebox{\textwidth}{!}{
  \begin{tabular}{lclllll}
    \toprule\small
    \sc authors & \sc at &\sc host & \sc size &  \sc size unit &\sc domain & \sc license\\\midrule
    \citet{papantoniou2023automating} & \color{red} A & \href{https://zenodo.org/records/7429037}{Zenodo} & 18,615 & \sc doc & \sc events &  \sc \href{https://creativecommons.org/licenses/by/3.0/legalcode}{cc by 3.0} \\
    \citet{koniaris2023evaluation} & \emph{C} & \href{https://huggingface.co/datasets/DominusTea/GreekLegalSum}{Hugging Face} & 8,395 & \sc doc & \sc legal &  \sc \href{https://creativecommons.org/licenses/by-nc/4.0/deed.en}{cc by-nc 4.0} \\
    \citet{rizou2023efficient} & \emph{M} & \href{https://msensis.com/wp-content/uploads/2023/06/uniway.zip}{msensis.com} & 2,176 & \sc sent & \sc admin & \sc \href{https://creativecommons.org/licenses/by-nc-sa/4.0/deed.en}{cc-by-nc-sa 4.0} \\
    \citet{dritsa2022greek} & \color{red} A & \href{https://zenodo.org/records/7005201}{Zenodo} & 1.28M & \sc doc &\sc politic &  \sc \href{https://creativecommons.org/licenses/by/4.0/deed.en}{cc by 4.0}\\
    \citet{papaloukas2021} &\emph{C} & \href{https://huggingface.co/datasets/greek_legal_code}{Hugging Face}& 47,563 & \sc doc & \sc legal &  \sc \href{https://creativecommons.org/licenses/by/4.0/deed.en}{cc by 4.0}\\
    \citet{korre2021elerrant} & \emph{M} & \href{https://github.com/katkorre/elerrant}{GitHub}& 227& \sc sent & \sc essays & \sc \href{https://opensource.org/license/mit}{mit}\\
    \citet{Zampieri2020} & \emph{M} & \href{https://huggingface.co/datasets/strombergnlp/offenseval_2020/viewer/gr/}{Hugging Face}& 10,287 & \sc tweet &\sc general & \sc \href{https://creativecommons.org/licenses/by/4.0/deed.en}{cc by 4.0}\\
    \citet{Bartziokas2020} & \emph{H} & \href{https://github.com/nmpartzio/elNER}{GitHub}& 21,153& \sc sent & \sc news &\sc \href{https://creativecommons.org/licenses/by-nc-sa/4.0/deed.en}{cc-by-nc-sa 4.0}\\
    \citet{prokopidis2020neural} & \color{red} N & \href{http://nlp.ilsp.gr/setn-2020/}{nlp.ilsp.gr} & 101,857 & \sc doc &\sc web & \sc \href{https://creativecommons.org/licenses/by-nc-sa/4.0/deed.en}{cc-by-nc-sa 4.0}\\
    \citet{Barzokas2020} & \emph{C} & \href{https://github.com/intelligence-csd-auth-gr/greek-words-evolution.git}{GitHub} & 1,734 & \sc doc & \sc e-books & \sc \href{https://opensource.org/license/mit}{mit}\\
    \citet{prokopidis2017universal} & \emph{H} & \href{https://github.com/UniversalDependencies/UD_Greek-GDT}{GitHub} & 2,521 & \sc sent & \sc news, politic &  \sc \href{http://creativecommons.org/licenses/by-nc-sa/3.0/}{cc by-nc-sa 3.0} \\
    \citet{prokopidis-2016-parallel} & \emph{U} & \href{http://nlp.ilsp.gr/pgv/}{nlp.ilsp.gr}& 17,018 & \sc doc pair & \sc news &  \sc \href{https://creativecommons.org/licenses/by/4.0/deed.en}{cc by 4.0}\\
    \bottomrule
  \end{tabular}}
  \caption{Greek datasets with their annotation type (AT) defined as manual (M), automatic (A), curated (C; metadata provided by the distributor), hybrid (H; manual and automatic), user-generated (U; from user-edits, not curated) or no (N). We include also the host, size, size unit, domain and license of each dataset.}
  \label{tab:data}
  \vspace{-10pt}
\end{table*}

\subsection{Datasets with no supervision signal}\label{ssec:raw}
Table~\ref{tab:data} comprises three datasets with no ground truth or whose ground truth is automatically extracted ({\color{red}A/N}; i.e., of doubted quality). Although excluded from our benchmarking, these datasets may still serve pre-training purposes and are further discussed next. \citet{dritsa2022greek} gathered 1.28M political speeches spanning from July 1989 to July 2020. \citet{papantoniou2023automating} conducted NER and entity linking on a dataset derived from Greek Wikipedia event pages. \citet{prokopidis2020neural} provided 101,857 files, which comprise the cleaned version of websites containing open content, including online archives of Greek newspapers from 2003 to 2020, and the Greek part of the W2C corpus \cite{majlivs2012language}. 

\subsubsection{Dataset statistics} 
\begin{table}[t]
  \centering\resizebox{.5\textwidth}{!}{
    \begin{tabular}{lllll}
    \toprule
         &\sc min &\sc max &\sc avg &\sc size (\#) \\
         \midrule
         \citet{papantoniou2023automating} & 34 & 1,484 & 172.9 & 18,615 \\ 
         \citet{prokopidis2020neural} & 36 & 67,963 & 2,430 & 101,857 \\ 
         \citet{dritsa2022greek} & 1 & 291,949 & 900 & 1,280,927 \\
    \bottomrule
    \end{tabular}
    }
    \caption{Greek FAIR datasets appropriate for pre-training tasks with min, max, average length in characters and size in documents.}
    \label{tab:raw}\vspace{-.5cm}
\end{table}
As is shown in Table~\ref{tab:raw}, the text length (measured in characters) greatly varies between the datasets. The dataset with the shortest texts on average is that of \citet{papantoniou2023automating} and that with the lengthiest is that of \citet{prokopidis2020neural}. The dataset size (counted in number of files) also varies, starting from ten thousands \cite{papantoniou2023automating}, to one hundred thousands \cite{prokopidis2020neural}, to more than a million \cite{dritsa2022greek}. The aggregated text totals up to 1.4 billion characters and 211 million tokens (using white space split).

\begin{figure}[t]
    \centering
    \includegraphics[width=.5\textwidth]{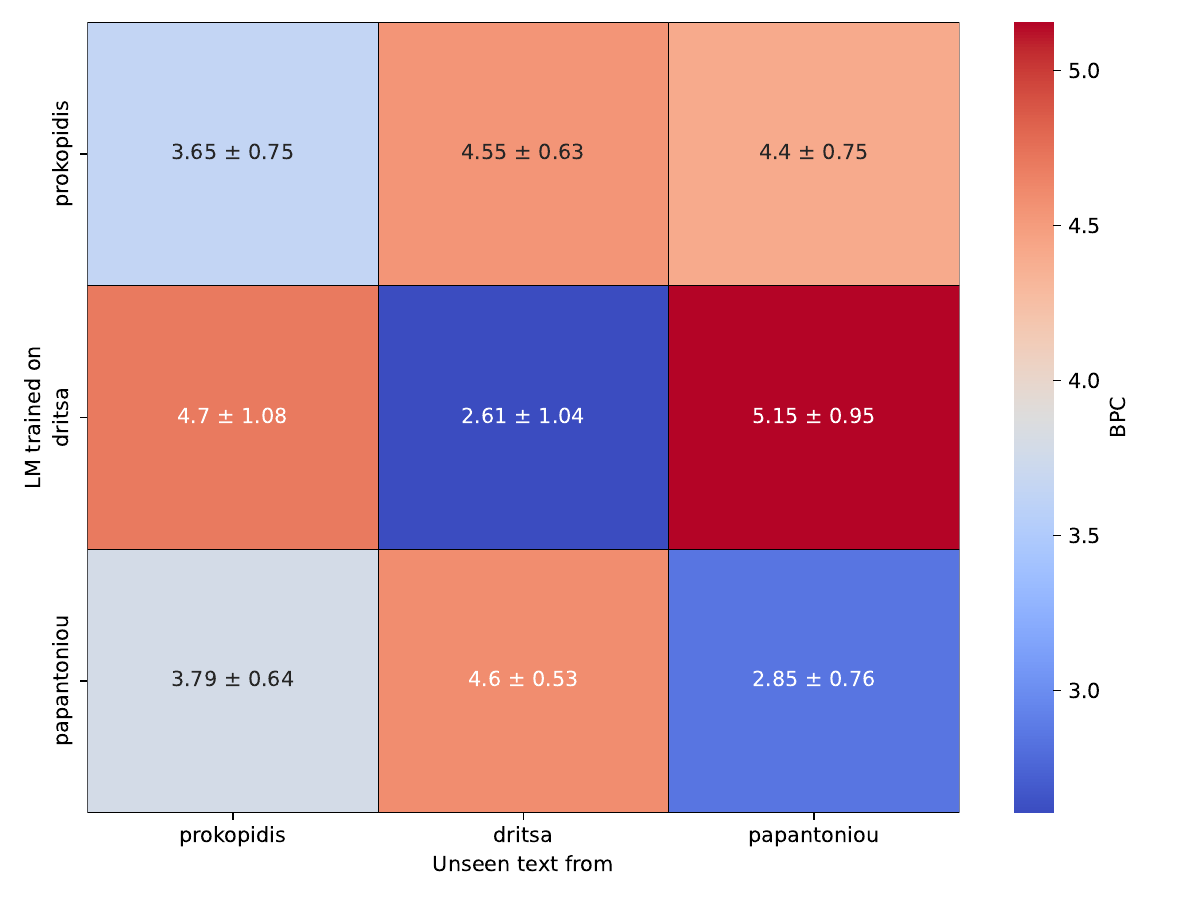}
    \caption{Heatmap of BPC measured on a sample per dataset (average across the sample; shown horizontally) per trained LM (vertically). Distance in warm colours.}
    \label{fig:ppl_heatmap}
\vspace{-15pt}
\end{figure}

\subsubsection{Linguistic distance}
We investigate the linguistic associations between these datasets, using statistical language modelling (LMing) as a proxy. For each dataset, we trained one statistical character-level LM on the first 100 characters of 1,000 randomly selected texts per dataset.\footnote{Character-level statistical LMing is an efficient way to capture stylistic features, employed in authorship analysis \cite{pavlopoulos2023computational}.} Then, we computed the Bits per Characters (BPC) across 500 randomly selected texts per dataset, reporting the mean. The result is displayed as a heatmap in Figure~\ref{fig:ppl_heatmap}. High scores (in red) reflecting a greater linguistic distance between the dataset the LM is trained on and the one BPC is computed on. Low scores (in blue) reflect a low linguistic distance, with the lowest scores taking place in the diagonal, when predicting unseen text taken from the same dataset (self-distance). Based on this heatmap, we find that the dataset of \citet{prokopidis2020neural} shows a higher self-distance compared to the rest while the datasets of \citet{dritsa2022greek} and \citet{papantoniou2023automating} are far away from each other, especially when the LM is trained on the former and predicting the latter.  

\subsection{Datasets with supervision signal}
The datasets that can be used as benchmarks for downstream tasks in Greek are presented in Table~\ref{tab:downstream_stats}. They cover supervised tasks (i.e., Intent, Toxicity, Authorship) and unsupervised tasks (i.e., Clustering) learning; text-to-text tasks (i.e., MT, GEC, Summarization); and sequence learning tasks (i.e., NER, POS tagging). The task with the lengthiest texts is Clustering and the one with the shortest is POS tagging. We chose Clustering over legal text classification because the dataset serves as a valuable resource for Clustering, a task currently lacking a benchmark. The dataset in this case has 47 clusters (classes), and it also comprises more fine-grained labels, increasing this number to 374 and 1,685. The task with the highest number of classes besides Clustering is NER (18), followed by Authorship (17) and POS tagging (16). Intent classification is the only balanced task ($CIR=1.0$).  

\begin{table*}[t]
    \centering\resizebox{\textwidth}{!}{
    \begin{tabular}{lcllllllll}
    \toprule
          \sc task & \sc dataset & \sc min &\sc max & \sc avg & \sc test size & \sc unit & \sc task & \sc classes \# & \sc cir \\\midrule
         \sc toxicity & \citet{zampieri2019predicting} & 8 & 289 & 107 & 1,544 & sent & \sc clf & 2 & 0.36\\
         \sc authorship & \citet{Barzokas2020} & 1,000, & 1,000 & 1,000 & 179 $\dagger$& text & \sc clf & 17 & 0.18 \\
         \sc intent & \citet{rizou2023efficient} & 7 & 174 & 59 & 436 $\dagger$& sent & \sc clf & 6 & 1.0 \\
         \sc clustering & \citet{papaloukas2021} & 17 & 931,736 & 42 & 9,516 & text & \sc clf & 47 & 0.07\\
         \sc gec & \citet{korre2021elerrant} & 19 & 482 & 134 & 227 & sent & \sc t2t & -- & -- \\
         \sc mt & \citet{prokopidis-2016-parallel} & 2 & 976 & 132 & 1,500$\dagger$ &doc & \sc t2t & -- & -- \\       
         \sc summarization & \citet{koniaris2023evaluation} & 2,571 & 6,983 & 4,899 & 1,238 & text & \sc t2t & -- & -- \\
         \sc ner & \citet{Bartziokas2020} & 2 & 2,210 & 175 & 299 $\dagger$ & sent & \sc sl & 4/18 & 0.02 \\
         \sc pos & \citet{prokopidis2017universal} & 1 & 96 & 24 & 456 & sent & \sc sl & 16 & 0.03 \\
         \bottomrule
    \end{tabular}}
    \caption{Statistics (counting in chars) for datasets related to downstream NLP tasks in Greek: classification (\textsc{clf}), sequence labelling (\textsc{sl}), text to text (\textsc{t2t}). When applicable, the class imbalance ratio (CIR: the smallest to the largest class) is reported. Datasets split by us are marked with a dagger. }
    \label{tab:downstream_stats}\vspace{-10pt}
\end{table*}

\section{The Greek NLP Benchmark}
We used licensed, accessible, machine-actionable, and with no hidden costs datasets, summarised in Table \ref{tab:downstream_stats}, to set up a benchmark per task. The ground truth is derived either by human annotators or by metadata sourced from the distributors. We experimented with 0-shot learning using \llama (70b-instruction)\footnote{We also experimented with a 4-bit quantized version of \href{https://huggingface.co/ilsp/Meltemi-7B-Instruct-v1-AWQ}{Meltemi}, a Mistral-7B Instruct model that was continually pre-trained on Greek \cite{voukoutis2024meltemi}, but the model failed to understand the task at hand (i.e., Toxicity and GEC).} and \gpt (4o-mini).\footnote{In the sequence learning tasks of POS tagging and NER, each class/tag was provided along with explanations.} We used the test splits, when they were available,\footnote{We created and provide our own splits, where these were not available; marked with a dagger in Table~\ref{tab:downstream_stats}.} and we randomly sampled 175 instances per test set otherwise, mainly to limit the cost for \gpt.

\subsection{Toxicity Detection} 
\paragraph{The dataset} In 2020, a subtask of offensive language identification for Greek was introduced as part of the SemEval-2020 Task 12 on Multilingual Offensive Language Identification in Social Media (OffensEval-2020)\footnote{\url{https://sites.google.com/site/offensevalsharedtask/offenseval-2020}}~\cite{Zampieri2020}. The task focused on four other languages besides Greek: Danish, English, Turkish, and Arabic. Overall, 145 teams submitted official runs on the test data, 37 of which made an official submission on Greek, while the submissions for English were approx. the double, i.e. 81. The Greek dataset used for the SemEval subtask is an extended version of the Offensive Greek Tweet Dataset (OGTD), which was developed by \citet{Pitenis2020}. As can be seen in Table~\ref{tab:downstream_stats}, texts in this dataset are relatively short and there is only slight class imbalance. 

\paragraph{The results} Table~\ref{tab:results_toxicity} shows that \llama is better in F1 for the better-supported \textsc{not toxic} class (due to its better recall) and worse for the minor \textsc{toxic} class. Hence, \gpt appears superior in macro-averaged F1 but the two are equal in performance in weighted-F1.  
\begin{table}[ht]
\centering\small
\begin{tabular}{lccc}
\toprule
& \multicolumn{2}{c}{\textbf{F1-Score}} & \textbf{Support} \\ 
\cmidrule(lr){2-3}
& \llama & \gpt & \\ 
\midrule
\textsc{not toxic} & \bf 0.82 & 0.80 & 148 \\ 
\textsc{toxic} & 0.37 & \bf 0.50 & 27 \\ 
\midrule
\textbf{Macro Avg} & 0.60 & 0.65 & 175 \\ 
\textbf{Weighted Avg} & 0.75 & 0.75 & 175 \\ 
\bottomrule
\end{tabular}
\caption{F1 in Toxicity using (0-shot) \llama and \gpt. Precision and Recall shown in the Appendix (Table~\ref{tab:apx:results_toxicity}).}
\label{tab:results_toxicity}
\vspace{-10pt}
\end{table}

\subsection{Grammatical Error Correction} 
\paragraph{The dataset} This task concerns the correction of grammatical errors that vary from grammatical mistakes to  punctuation, spelling, and morphology of word. \citet{korre2021elerrant} listed 18 main categories of grammatical errors that systems can correct. They developed two datasets, of which we consider only the one that is annotated by human experts, i.e., the Greek Native Corpus (GNC). This corpus is comprised of essays written by students who are native speakers of Greek, totalling 227 sentences. Each sentence within this dataset may contain zero, one, or multiple grammatical errors, all annotated by human experts with the corresponding grammatical error types as defined in the provided annotation schema.

\paragraph{The results} Using character (CER) and word (WER) error rate, we find that \gpt performs significantly better than \llama in correcting grammatical errors. It is halved in WER compared to \llama and only 1.74 in CER. Unless \gpt has already used the data of \citet{korre2021elerrant} during training, a possibility we cannot exclude for this dataset, this is a very low rate. To gain a better insight, we experimented by also adding two shots in the prompts. Although the performance of \llama halved, that of \gpt remained unchanged. 
\begin{table}[ht]
\centering\small
\begin{tabular}{p{2cm}cc|cc}
\toprule
& \llama & \small -2s & \gpt &\small -2s \\ 
\midrule
\textsc{wer} & 14.51 &\small (-38\%) & \bf 7.55 & \small (-1\%) \\ 
\textsc{cer} & 8.69 & \small (-46\%) & \bf 1.74 & \small (+18\%) \\ 
\bottomrule
\end{tabular}
\caption{Word (WER) and character (CER) error rate of \llama and \gpt using 0-shot learning and the percentage change when adding 2 (static) shots.}
\label{tab:results_gec}
\vspace{-10pt}
\end{table}

\subsection{Machine Translation}\label{ssec:mt}
\paragraph{The dataset} \citet{prokopidis-2016-parallel} created bilingual corpora (756 language pairs) from content available in \href{https://globalvoices.org/}{Global Voices}, where volunteers translate news stories in 41 languages. The 3,629 Greek documents are translated into 40 languages. However, not every document is translated into all languages, resulting in 17,018 bilingual document pairs involving Greek. In this work, we focus on three Greek language pairs: Greek-English, Greek-Japanese, and Greek-Farsi. These target languages were selected as the most supported languages within their respective language support tiers, as defined in \citet{bakagianni2024towards}. The tiers are based on the number of ACL Anthology studies (2012-2024) referencing (in their titles/abstracts) each respective language.\footnote{As listed in the \href{https://www.ietf.org/rfc/bcp/bcp47.html}{IETF BCP 47 standard}.} Well-supported languages are referenced in more than 1,000 studies; moderately-supported languages are referenced in 100 to 1,000; less-supported ones are referenced in less than 100. Based on this classification, the English language represents the well-supported tier (6,915 studies), Japanese represents the moderately-supported tier (808 studies) and Farsi the less-supported tier (98 studies). We use this dataset to benchmark machine translation, experimenting with Greek as the source language.

\paragraph{The results} 
Table~\ref{tab:results_mt} presents the BERTScore F1 \cite{zhang2019bertscore} per model, along with CER and WER. 
\begin{table}[h]
\centering\small
\begin{tabular}{p{2cm}cc}
\toprule
& \llama & \gpt \\ 
\midrule
\multicolumn{3}{l}{\textbf{\textsc{el-en}: Greek to English}} \\ \midrule
WER & 65.53 (41.60) & \bf 53.13 (33.09) \\ 
CER & 48.68 (31.91) & \bf 36.63 (24.25) \\ 
BERTScore F1 & 0.80 & \bf 0.81 \\ 
\midrule
\multicolumn{3}{l}{\textbf{\textsc{el-jpn}: Greek to Japanese}} \\ \midrule
WER & 267.57 (541.60) & \bf 109.05 (41.13) \\ 
CER & 153.82 (194.99) & \bf 129.69 (186.95) \\ 
BERTScore F1 & 0.49 & \bf 0.54 \\ 
\midrule
\multicolumn{3}{l}{\textbf{\textsc{el-fa}: Greek to Farsi}} \\ \midrule
WER & 92.67 (38.83) & \bf 88.81 (34.41) \\ 
CER & 71.01 (31.07) & \bf 67.00 (32.70) \\ 
BERTScore F1 & \bf 0.54 & \bf 0.54 \\ 
\bottomrule
\end{tabular}
\caption{The average (standard deviation) error rate (WER, CER) is computed across the three source-target language pairs, along with the BERTScore F1. We use \llama and \gpt 0-shot learning, and we report the best score achieved when multiple gold translations are available.}
\label{tab:results_mt}
\vspace{-10pt}
\end{table}
By translating to English, a well-supported language, we report the best results across all metrics. By translating to Farsi, a less-supported language, we observe the worst performance, slightly below those for translations to Japanese, a moderately-supported language. Across all languages, \gpt demonstrates a lower translation error rate, in terms of WER and CER, compared to \llama, but its (BERTScore) F1 is only higher for Japanese. For English and Farsi, both models achieve similar BERTScore F1. This suggests that while \llama produces translations of comparable quality, it tends to use different wording than the ground truth. Overall, we observe that the translation quality from Greek to the selected languages reflects their classification within the language support tiers.

\subsection{Summarization}\label{ssec:summarization}
\paragraph{The dataset} \citet{koniaris2023evaluation} created a legal corpus of 8,395 court decisions from Areios Pagos, the Supreme Civil and Criminal Court of Greece. This corpus includes the decisions, their summaries, and related metadata, all sourced from the Areios Pagos website.\footnote{\url{https://www.areiospagos.gr/}} Using the (provided) test set of 1,238 Greek legal texts, we find an average text length of 18,541 characters and a standard deviation of 18,351 (min: 2,571 and max: 303,538). When tokenised (white-space split), this is 381 (min) to 46,232 (max) tokens. We only kept texts of 1,000 tokens or fewer, to limit the time and cost of our experiments. This totals 192 texts of 4,899 characters on average with a standard deviation of 1,319 characters (max: 6,983). To remain consistent with the benchmarking of other NLP tasks, we sampled 175 texts for the experiments.

\paragraph{The results} In addition to BERTScore \cite{zhang2019bertscore}, we also report variants of ROUGE \cite{lin-2004-rouge}. As is shown in Table~\ref{tab:results_summarization}, \llama outperforms \gpt across most metrics for this task, demonstrating stronger overall performance in generating high-quality summaries. However, \gpt slightly surpasses \llama in recall, as measured by ROUGE (across variants), suggesting that it better captures content overlap, though at the expense of precision.

\begin{table}[ht]
\centering\small
\begin{tabular}{lcc}
\toprule
\textbf{Metric} & \llama & \gpt \\ 
\midrule
BERTScore & \textbf{0.517 ± 0.108} & 0.509 ± 0.101 \\ 
ROUGE-1 & \textbf{0.192 ± 0.089} & 0.186 ± 0.076 \\ 
ROUGE-2 & \textbf{0.048 ± 0.055} & 0.039 ± 0.037 \\ 
ROUGE-L & \textbf{0.167 ± 0.080} & 0.164 ± 0.066 \\ 
\bottomrule
\end{tabular}
\caption{Mean (±SD) BERTScore and ROUGE F1 of \llama and \gpt. More metrics are shown in Table~\ref{tab:apx:results_summarization}.}
\label{tab:results_summarization}
\end{table}\vspace{-15pt}

\subsection{Intent Classification}
\paragraph{The dataset} \citet{rizou2023efficient} collected student queries to two University help desks and manually annotated each of these with three entity tags and six intents. The dataset is balanced regarding the intents. The average text length in characters is 58.6, the maximum is 174 (minimum of 7), and the standard deviation is 30.3. 

\paragraph{The results} As shown in Table~\ref{tab:results_intent}, \gpt is overall (F1 of 0.93) and consistently outperforms \llama (F1 of 0.70). \llama faces more challenges with Recall than with Precision. In certain classes, the Recall score for \llama was as low as 0.24, which significantly reduces its macro-average to 0.72.  
\begin{table}[ht]
\centering\small
\begin{tabular}{lccc}
\toprule
\textbf{Intent} & \textbf{\llama} & \textbf{\gpt} & \textbf{Support} \\ 
\midrule
\tiny \textsc{availableCoursesbyExamPeriod} & 0.86 & \textbf{0.91} & 30 \\ 
\tiny\textsc{coefficientsByCourseName} & 0.39 & \textbf{0.98} & 29 \\ 
\tiny\textsc{gradeByCourseName} & 0.85 & \textbf{0.93} & 29 \\ 
\tiny\textsc{passedCoursesByExamPeriod} & 0.80 & \textbf{0.93} & 29 \\ 
\tiny\textsc{teacherInfoByName} & 0.69 & \textbf{0.92} & 29 \\ 
\tiny\textsc{teacherNameByCourseName} & 0.60 & \textbf{0.89} & 29 \\ 
\midrule
\textbf{Macro Avg} & 0.70 & \textbf{0.93} & 175 \\ 
\textbf{Weighted Avg} & 0.70 & \textbf{0.93} & 175 \\ 
\bottomrule
\end{tabular}
\caption{F1 in intent using (0-shot) \llama and \gpt. Precision and Recall shown in the Appendix (Table~\ref{tab:apx:results_intent}).}
\label{tab:results_intent}
\vspace{-10pt}
\end{table}

\subsection{NER}

\paragraph{The dataset} \citet{Bartziokas2020} provided an annotated dataset with two levels of granularity for entity annotation. The first level uses 4 label tags akin to the CONLL-2003 dataset \cite{sang2003introduction}, while the second incorporates 18 entity tags, as in the OntoNotes 5 English dataset \cite{pradhan2007ontonotes}. The dataset was developed during a Google Summer of Code project in 2018,\footnote{\url{https://github.com/eellak/gsoc2018-spacy}}, where initial automatic annotation was followed by manual curation. The text length varies from 2 to 2,210 characters (Table~\ref{tab:downstream_stats}). Both annotation levels exhibit heavy imbalance, with the four-tag level having an imbalance of 0.02 and the 18-tag level showing near-zero imbalance. We opted for the four-entity-tag level of annotation with the lower imbalance. 

\paragraph{The results} 
Table~\ref{tab:results_ner} shows that \llama achieves a better F1 score overall compared to \gpt. However, the low macro-average scores indicate that both models fail to capture specific entity classes, such as E-MISC and S-MISC (see Appendix (Table~\ref{tab:apx:results_ner})). Both models perform well in detecting the O class (i.e., not an entity), yielding an F1 of 0.92, which means that both models can detect entities. The low macro average scores, on the other hand, reflect how both models struggle to distinguish between the different types of entities.   
\begin{table}[h!]
\centering\small
\begin{tabular}{lccc}
\toprule
\textbf{Metric} & \textbf{\llama} & \textbf{\gpt} & \textbf{Support} \\ 
\midrule
Macro Avg & \bf 0.14 & \bf 0.14 & 5447 \\ 
Weighted Avg & 0.85 & \bf 0.84 & 5447 \\ 
\bottomrule
\end{tabular}
\caption{F1 in NER using (0-shot) \llama and \gpt. Results per entity tag shown in the Appendix (Table~\ref{tab:apx:results_ner}).}
\label{tab:results_ner}
\vspace{-20pt}
\end{table}

\subsection{POS Tagging}

\paragraph{The dataset}
\citet{prokopidis2017universal} provided the Greek treebank as part of a project that offers standardized treebanks with consistent annotations across languages \cite{nivre2016universal}. The dataset includes syntactic dependencies, POS tags, morphological features, and lemmas. It contains 2,521 sentences split into train (1,622), development (403), and test (456) sets, and was manually validated and corrected.

\paragraph{The results}
Table~\ref{tab:results_pos} shows that \gpt is consistently better than \llama. For the POS tags DET (Determiner) and VERB (Verb), \llama achieves higher recall, meaning it identifies more instances of these tags (see Appendix (Table~\ref{tab:apx:results_pos})). However, \gpt has a better overall F1 score, indicating a better balance between precision and recall. Neither model is able to correctly handle unknown or missing tags (represented as \_) or other foreign words, abbreviations, etc. (denoted as X).
\begin{table}[ht]
\centering\small
\begin{tabular}{lccc}
\toprule
\textbf{Metric} & \textbf{\llama} & \textbf{\gpt} & \textbf{Support} \\ 
\midrule
Macro Avg & 0.35 & \textbf{0.47} & 4131 \\ 
Weighted Avg & 0.48 & \textbf{0.60} & 4131 \\ 
\bottomrule
\end{tabular}
\caption{F1 in POS tagging using (0-shot) \llama and \gpt. Results per tag shown in the Appendix (Table~\ref{tab:apx:results_pos}).}
\label{tab:results_pos}
\vspace{-10pt}
\end{table}

\section{Novel Benchmark Tasks for Greek}

\subsection{Long Legal Text Clustering}\label{ssec:clustering}
\paragraph{The dataset} \citet{papaloukas2021} developed a dataset for multi-class legal topic classification, derived from a collection of Greek legislative documents titled ``Permanent Greek Legislation Code - \href{https://raptarchis.gov.gr}{Raptarchis}''. The dataset includes classifications that range from broader categories to more specialized ones. There are annotations at the volume, the chapter and the subject levels. This is a heavily imbalanced dataset, and the text lengths vary significantly, from 17 to approx. 1 million characters. The high number of classes in this task make in-context learning approaches impractical while our benchmark already covers text classification as a task (Table~\ref{tab:downstream_stats}). Therefore, we use this dataset to benchmark LLMs for text clustering, a task of unsupervised learning that is currently missing for Greek data in literature. We employ the entire test set of 38,052 texts for our experiments, comprising 47 volumes, 374 chapters, and 1,685 subjects. 

\paragraph{The results} The direct use of LLM models for this task is prohibitive due to the substantial computational cost and processing time required for handling the long legal texts (Table~\ref{tab:downstream_stats}). Therefore, we opted for K-Means with TF-IDF features.\footnote{We used \href{https://scikit-learn.org/}{scikit-learn} with default parameters.} Additionally, we used \llama for summarization (the best-performing model for this task; see \S\ref{ssec:summarization}) and for translation into English (performing on par with \gpt; see \S\ref{ssec:mt}). This allowed us to compute Instructor embeddings \cite{su2022one}, which we refer to as STE (Summarized, Translated, Embedded), as an alternative to TF-IDF for text representation. We set the number of clusters, $k$, according to the ground truth number of topics, i.e., 47 for the volume, 374 for the chapter, and 1,685 for the subject. We evaluated the results based on the normalised mutual information score (NMI), the adjusted mutual information score (AMI), and the adjusted rand index (ARI). All three measures use the ground truth labels to assess the clustering solution while being independent of the absolute label values. As shown in Table~\ref{tab:results_clustering}, STE (2nd row per level) gives the best results overall, and the best ones across metrics consistently in two levels. When ablating the embedding step, computing the TF-IDF features of the English summaries (3rd row per level), we see that the results deteriorate. This means that information is lost during translation and summarization. The superiority of the Instructor embeddings, however, make up for this drop.
\begin{table}[h!]
    \centering\small
    \begin{tabular}{ccccccc}\toprule
    \multirow{2}{*}{\bf $k$} & \multirow{2}{*}{\bf \textsc{rep}} & \multirow{2}{*}{\sc \textbf{lan}} & \multirow{2}{*}{\sc \textbf{len}} & \multicolumn{3}{c}{\sc \textbf{metrics}} \\\cmidrule{5-7}
    & & & & \sc nmi & \sc ami & \sc acc\\\midrule
    \multirow{3}{*}{47} & TF-IDF & Gr & Full & 0.161 & 0.133 & 0.124 \\
    & Dense & En & Sum & \bf 0.244 &\bf 0.219 & \bf 0.194 \\
    & TF-IDF & En & Sum & 0.136 & 0.121 & 0.129 \\
    \midrule
    \multirow{3}{*}{374} & TF-IDF & Gr & Full & 0.479 & 0.194 & 0.082 \\
    & Dense & En & Sum & \bf 0.524 &\bf 0.225 & \bf 0.175 \\
    &TF-IDF & En & Sum & 0.321 & 0.156 & 0.118 \\
    \midrule
    \multirow{3}{*}{1,685} & TF-IDF & Gr & Full & 0.723 & \bf 0.202 & 0.057 \\
    & Dense & En & Sum & \bf 0.725 & 0.162 & \bf 0.246 \\
    & TF-IDF & En & Sum & 0.510 & 0.111 & 0.162 \\
    \bottomrule
    \end{tabular}
    \caption{Clustering evaluation for long legal Greek documents \cite{papaloukas2021}, comparing KMeans and STE (Dense-En-Sum) at three granularity levels ($k=47$, $k=374$, $k=1,685$).}
    \label{tab:results_clustering}
\end{table}
\vspace{-15pt}

\subsection{Authorship Attribution}\label{ssec:authorship}
\paragraph{The dataset} \citet{Barzokas2020} collected 1,734 open-access e-books from Project Gutenberg,\footnote{\url{https://www.gutenberg.org/}} and from Open Library. Due to the conversion from various format to plain text, the 1,105 e-books have content in plain text that corresponds to the original file content. For our experiments, we filtered out books with more than 1,000 tokens (the number of tokens was reported in the dataset). From the remaining 985 books, to yield a test set (of 175 instances, as in other tasks), we selected books from authors with many books, resulting in 175 books from 17 authors. For each book, we used 1,000-character excerpts taken from the middle of the text to detect the author of a given passage.

\paragraph{The results} Table~\ref{tab:results_authorship} shows the results of \llama and \gpt, with the latter performing overall better. \llama outperforms \gpt for three authors - Tzouvalis, Amanatidou, and Aeschylus - in terms of F1-score, with both models achieving perfect Precision for these authors, suggesting prior exposure to their works during the pre-training phase. Similar indications arise for Papavasileiou and Plato, who achieve the best results for both models. However, \gpt achieved a zero F1-score for Voulazeris, Kritsotaki, Semertzidou, Fourouklas, and Kappas, indicating that data leakage or contamination is unlikely for these authors, as the models fail to correctly classify their works. \llama also achieved a zero F1-score for these authors, as well as for Koliopoulos, Andamis, and Zervas. 
\begin{table}[h!]
\centering\small
\begin{tabular}{lccc}
\toprule
\textbf{Author} & \textbf{\llama} & \textbf{\gpt} & \textbf{Support} \\ 
\midrule
Thanasis Triaridis & 0.32 & \textbf{0.34} & 28 \\ 
Rania Synodinou & 0.11 & \textbf{0.41} & 12 \\ 
Kostas Voulazeris & 0.00 & 0.00 & 17 \\ 
Dimitris Tzouvalis & \textbf{0.47} & 0.33 & 13 \\ 
Evridiki Amanatidou & \textbf{0.40} & 0.46 & 7 \\ 
Frinta Kritsotaki & 0.00 & 0.00 & 5 \\ 
Panos Koliopoulos & 0.00 & \textbf{0.10} & 17 \\ 
Yiannis Andamis & 0.00 & \textbf{0.06} & 23 \\ 
Giorgos S. Kokkinos & 0.29 & \textbf{0.42} & 8 \\ 
Manos Kounougakis & 0.25 & \textbf{0.33} & 5 \\ 
Eleni Semertzidou & 0.00 & 0.00 & 5 \\ 
Panos A. Zervas & 0.00 & \textbf{0.15} & 7 \\ 
Lakis Fourouklas & 0.00 & 0.00 & 5 \\ 
Vasileios Kappas & 0.00 & 0.00 & 5 \\ 
Paschalis Papavasileiou & 0.80 & \textbf{0.89} & 5 \\ 
Plato & 0.75 & \textbf{0.82} & 8 \\ 
Aeschylus & \textbf{0.57} & 0.33 & 5 \\ 
\midrule
\textbf{Macro Avg} & 0.23 & \textbf{0.27} & 175 \\ 
\textbf{Weighted Avg} & 0.20 & \textbf{0.25} & 175 \\ 
\bottomrule
\end{tabular}
\label{tab:results_authorship}
\caption{F1 in Authorship with (0-shot) \llama and \gpt. Precision and Recall in Table~\ref{tab:apx:results_authorship} of the Appendix.}
\end{table}
\vspace{-20pt}

\section{Related work}

Our study provides a unified cross-task Greek NLP benchmark, based on pre-existing datasets. Our work is not the first to reveal existing datasets. \cite{Nikiforos2021}, for example, reviewed methods and datasets related to the Greek social web. \citet{Alexandridis2021Survey} focused on sentiment analysis in Greek social media while \citet{Krasadakis2021} surveyed NLP studies related to legislative and Greek documents. These studies, however, have not assessed the employed datasets regarding their licensing, public availability, and (machine) actionability, which hinders the impact of existing benchmarks. We provide an extensive benchmark on publicly available and actionable Greek NLP data, hence filling this gap. 

Overviews that focus on Greek NLP, such as the work of \citet{papantoniou2020nlp,papantoniou2024nlp}, cover a very wide period of study (e.g., comprising Ancient Greek), and lack classification for licensing and actionability of the resources. The systematic literature review of \citet{bakagianni2024towards} followed that path and provided an exhaustive list of publicly available and actionable datasets. Our work was based on that study, providing code to compile a unified collection of appropriate datasets, to ease usability and re-production of the results. 

Data leakage and contamination is a challenge for the NLP community \cite{balloccu-etal-2024-leak}, which recent benchmarks employing LLMs attempt to avoid \cite{white2024livebench}. Benchmarks can assist towards that path, by providing specific evaluation sets and documentation. Although benchmarks exist for Greek dialects \cite{faisal2024dialectbench} and Ancient Greek \cite{stopponi2024agree}, this is the first benchmark across Greek NLP tasks using publicly available and actionable data.

\section{Conclusions}
Based on a monolingual survey, we compiled a collection of publicly available and accessible Greek datasets, based on their licensing schemas. We used this collection to benchmark an open- (\llama) v. a closed-source (\gpt) LLM on seven core NLP tasks, showing that the former best performs in NER and Summarisation while the latter best-performs in POS tagging, Intent Classification and GEC. We further observe and address two weaknesses in Greek NLP. First, the available data for authorship attribution are of limited availability. We tackled this by introducing an alternative dataset and by using 0-shot learning. High accuracy in this task could indicate usage of the respective material during pre-training. Second, there is no text clustering benchmark, which we addressed by introducing the first long legal text clustering benchmark. By summarising, translating and embedding we showed, then, that better representations can be resulted compared to TF-IDF representations. 

\section*{Acknowledgments}
This work has been partially supported by project MIS 5154714 of the National Recovery and Resilience Plan Greece 2.0 funded by the European Union under the NextGenerationEU Program.

\section{Limitations} 
We compiled a collection of existing datasets, providing the code to re-compile our collection. Dataset developers, however, may update their datasets, which may yield slightly altered results in one or more tasks. To reflect any such changes, we will regularly re-compile the collection and re-run our benchmark, reporting the results over time in our repository online. 
A limitation of our work lies in the sustainability of our code, to incorporate new repositories and adjust to changes in the current one. This is addressed by unit tests set to download periodically the data and evaluate the functionality. We also note that our method is applicable to other languages, if a systematic literature review for that language exists.   

\section{Ethical considerations}
Greek NLP datasets present a range of legal and licensing challenges primarily related to copyright, permissions, and data protection, crucial for compliant and ethical research use.

\paragraph{Copyright}
The dataset of \citet{papantoniou2023automating}, for example, relies on Greek Wikipedia, which may include third-party contributions under different licensing terms. This could lead to potential copyright issues if the dataset includes trademarked names or logos, not covered under the CC license and requiring separate permissions. \citet{Barzokas2020} created a dataset using content from Project Gutenberg and Open Library under an MIT license, but content of the former may not be free of copyright in other countries. Also, Open Library restricts its content use to non-commercial purposes and for research only, which may conflict with the broad permissions of the MIT license if the content is used beyond these limits.

\paragraph{Re-identification}
The judicial rulings of \citet{koniaris2023evaluation} may include copyrighted annotations by court staff, posing a risk of re-identification. Similarly, in the student essays of \citet{rizou2023efficient} there could be potential re-identification risks of student data (even if they are fully anonymised) while in the handwritten essays from high school students of \citet{korre2021elerrant} there may be concerns about obtaining proper consent from students or their guardians for dataset inclusion. On the other hand, the data of \citet{Bartziokas2020} may include named entities tied to individuals and there is potential for privacy violations. 

\paragraph{Conflicting terms} Derived from the Permanent Greek Legislation Code - Raptarchis, the dataset of \citet{papaloukas2021} could encounter legal concerns due to the Ministry of Digital Governance retaining all intellectual property rights. The dataset of \citet{dritsa2022greek} sourced from the Hellenic Parliament, conflicts with its public terms, which require proper attribution without alterations (e.g., 3rd party content). 
\citet{Zampieri2020} applied a CC-BY-4.0 license to tweets, presenting legal challenges due to Twitter’s restrictive policies on content redistribution. Also, regarding the data of \citet{prokopidis2020neural}, while derived from web content labeled as ``open'', the terms of the original websites must explicitly allow crawling and reuse to avoid potential copyright infringement.

\bibliography{references}

\appendix
\section*{Appendix}

\begin{table*}[ht]
\centering\small
\caption{F1 in Toxicity Detection with 0-shot \llama and \gpt.}
\begin{tabular}{lccccccc}
\toprule
& \multicolumn{2}{c}{\textbf{Precision}} & \multicolumn{2}{c}{\textbf{Recall}} & \multicolumn{2}{c}{\textbf{F1-Score}} & \textbf{Support} \\ 
\cmidrule(lr){2-3} \cmidrule(lr){4-5} \cmidrule(lr){6-7}
& \llama & \gpt & \llama & \gpt & \llama & \gpt & \\ 
\midrule
\textsc{not toxic} & 0.90 & 0.98 & 0.76 & 0.67 & 0.82 & 0.80 & 148 \\ 
\textsc{toxic} & 0.29 & 0.34 & 0.52 & 0.93 & 0.37 & 0.50 & 27 \\ 
\midrule
\textbf{Macro Avg} & 0.59 & 0.66 & 0.64 & 0.80 & 0.60 & 0.65 & 175 \\ 
\textbf{Weighted Avg} & 0.80 & 0.88 & 0.73 & 0.71 & 0.75 & 0.75 & 175 \\ 
\bottomrule
\end{tabular}
\label{tab:apx:results_toxicity}
\end{table*}

\begin{table}[ht]
\centering\small
\begin{tabular}{lcc}
\toprule
\textbf{Metric} & \llama & \gpt \\ 
\midrule
BERTScore F1 & \textbf{0.517 ± 0.108} & 0.509 ± 0.101 \\ 
BERTScore Precision & \textbf{0.441 ± 0.093} & 0.435 ± 0.087 \\ 
BERTScore Recall & \textbf{0.627 ± 0.132} & 0.616 ± 0.126 \\\hline
ROUGE-1 F1 & \textbf{0.192 ± 0.089} & 0.186 ± 0.076 \\ 
ROUGE-1 Precision & \textbf{0.185 ± 0.107} & 0.152 ± 0.083 \\ 
ROUGE-1 Recall & 0.249 ± 0.129 & \textbf{0.299 ± 0.129} \\\hline 
ROUGE-2 F1 & \textbf{0.048 ± 0.055} & 0.039 ± 0.037 \\ 
ROUGE-2 Precision & \textbf{0.046 ± 0.054} & 0.032 ± 0.031 \\ 
ROUGE-2 Recall & 0.069 ± 0.094 & \textbf{0.074 ± 0.103} \\\hline 
ROUGE-L F1 & \textbf{0.167 ± 0.080} & 0.164 ± 0.066 \\ 
ROUGE-L Precision & \textbf{0.159 ± 0.092} & 0.134 ± 0.071 \\ 
ROUGE-L Recall & 0.221 ± 0.129 & \textbf{0.269 ± 0.125} \\ 
\bottomrule
\end{tabular}
\caption{Mean (± SD) BERTScore and ROUGE F1, Precision, Recall of \llama and \gpt.}
\label{tab:apx:results_summarization}
\end{table}

\begin{table*}[ht]
\centering\small
\begin{tabular}{lccccccc}
\toprule
\textbf{Intent} & \multicolumn{2}{c}{\textbf{Precision}} & \multicolumn{2}{c}{\textbf{Recall}} & \multicolumn{2}{c}{\textbf{F1-Score}} & \textbf{Support} \\ 
\cmidrule(lr){2-3} \cmidrule(lr){4-5} \cmidrule(lr){6-7}
& \llama & \gpt & \llama & \gpt & \llama & \gpt & \\ 
\midrule
\sc getAvailableCoursesByExamPeriod & 0.75 & \textbf{0.83} & \textbf{1.00} & \bf 1.00 & 0.86 & \textbf{0.91} & 30 \\ 
\sc getCoeffByCourseName & \textbf{1.00} & \textbf{1.00} & 0.24 & \textbf{0.97} & 0.39 & \textbf{0.98} & 29 \\ 
\sc getGradeByCourseName & 0.83 & \textbf{0.93} & 0.86 & \textbf{0.93} & 0.85 & \textbf{0.93} & 29 \\ 
\sc getPassedCoursesByExamPeriod & 0.95 & \textbf{1.00} & 0.69 & \textbf{0.86} & 0.80 & \textbf{0.93} & 29 \\ 
\sc getTeacherInfoByTeacherName & 0.55 & \textbf{0.90} & \textbf{0.93} & \textbf{0.93} & 0.69 & \textbf{0.92} & 29 \\ 
\sc getTeacherNameByCourseName & 0.61 & \textbf{0.93} & 0.59 & \textbf{0.86} & 0.60 & \textbf{0.89} & 29 \\ 
\midrule
\textbf{Macro Avg} & 0.78 & \textbf{0.93} & 0.72 & \textbf{0.93} & 0.70 & \textbf{0.93} & 175 \\ 
\textbf{Weighted Avg} & 0.78 & \textbf{0.93} & 0.72 & \textbf{0.93} & 0.70 & \textbf{0.93} & 175 \\ 
\bottomrule
\end{tabular}
\caption{Evaluation of intent classification, comparing \llama{} and \gpt{} with 0-shot learning.}
\label{tab:apx:results_intent}
\end{table*}

\begin{table*}[h!]
\centering\small
\begin{tabular}{lccccccc}
\toprule
\textbf{NE} & \multicolumn{2}{c}{\textbf{Precision}} & \multicolumn{2}{c}{\textbf{Recall}} & \multicolumn{2}{c}{\textbf{F1-Score}} & \textbf{Support} \\ 
\cmidrule(lr){2-3} \cmidrule(lr){4-5} \cmidrule(lr){6-7}
& \llama & \gpt & \llama & \gpt & \llama & \gpt & \\ 
\midrule
S-LOC & \textbf{0.25} & 0.13 & \textbf{0.08} & 0.03 & \textbf{0.12} & 0.05 & 104 \\ 
B-LOC & \textbf{0.06} & 0.03 & \textbf{0.19} & 0.19 & \textbf{0.09} & 0.05 & 27 \\ 
I-LOC & 0.03 & \textbf{0.07} & 0.12 & \textbf{0.25} & 0.05 & \textbf{0.11} & 8 \\ 
E-LOC & \textbf{0.07} & 0.04 & \textbf{0.15} & 0.19 & \textbf{0.09} & 0.07 & 27 \\ 
S-ORG & \textbf{0.20} & 0.07 & \textbf{0.09} & 0.05 & \textbf{0.12} & 0.06 & 44 \\ 
B-ORG & 0.07 & \textbf{0.16} & 0.09 & \textbf{0.30} & 0.08 & \textbf{0.21} & 66 \\ 
I-ORG & 0.08 & \textbf{0.14} & 0.09 & \textbf{0.23} & 0.08 & \textbf{0.17} & 35 \\ 
E-ORG & 0.13 & \textbf{0.17} & 0.12 & \textbf{0.23} & 0.12 & \textbf{0.20} & 66 \\ 
S-PERSON & \textbf{0.07} & 0.04 & \textbf{0.08} & 0.08 & \textbf{0.08} & 0.06 & 12 \\ 
B-PERSON & \textbf{0.14} & 0.03 & \textbf{0.18} & 0.07 & \textbf{0.16} & 0.05 & 28 \\ 
I-PERSON & 0.08 & \textbf{0.04} & \textbf{1.00} & 1.00 & \textbf{0.15} & 0.07 & 1 \\ 
E-PERSON & 0.22 & \textbf{0.26} & 0.21 & \textbf{0.39} & 0.22 & \textbf{0.31} & 28 \\ 
S-MISC & 0.00 & 0.00 & 0.00 & 0.00 & 0.00 & 0.00 & 12 \\ 
B-MISC & \textbf{0.03} & 0.00 & \textbf{0.07} & 0.00 & \textbf{0.04} & 0.00 & 14 \\ 
I-MISC & \textbf{0.03} & 0.00 & \textbf{0.06} & 0.00 & \textbf{0.04} & 0.00 & 17 \\ 
E-MISC & 0.00 & 0.00 & 0.00 & 0.00 & 0.00 & 0.00 & 14 \\ 
O & 0.93 & \textbf{0.95} & \textbf{0.91} & 0.89 & \bf 0.92 & \bf 0.92 & 4944 \\ 
\midrule
\textbf{Macro Avg} & \textbf{0.14} & 0.13 & 0.20 & \textbf{0.23} & \bf 0.14 & \bf 0.14 & 5447 \\ 
\textbf{Weighted Avg} & 0.86 & \textbf{0.87} & \textbf{0.83} & 0.82 & \textbf{0.85} & 0.84 & 5447 \\ 
\bottomrule
\end{tabular}
\caption{Evaluation of NER per entity, comparing \llama{} and \gpt{} with 0-shot learning. The number of evaluation instances per class is shown rightmost.}
\label{tab:apx:results_ner}
\end{table*}

\begin{table*}[ht]
\centering\small
\begin{tabular}{lccccccc}
\toprule
\textbf{POS} & \multicolumn{2}{c}{\textbf{Precision}} & \multicolumn{2}{c}{\textbf{Recall}} & \multicolumn{2}{c}{\textbf{F1-Score}} & \textbf{Support} \\ 
\cmidrule(lr){2-3} \cmidrule(lr){4-5} \cmidrule(lr){6-7}
& \llama & \gpt & \llama & \gpt & \llama & \gpt & \\ 
\midrule
ADV & 0.45 & \textbf{0.57} & 0.29 & \textbf{0.54} & 0.35 & \textbf{0.55} & 164 \\ 
PUNCT & 0.57 & \textbf{0.70} & 0.55 & \textbf{0.67} & 0.56 & \textbf{0.69} & 406 \\ 
DET & 0.62 & \textbf{0.78} & \textbf{0.59} & 0.58 & 0.61 & \textbf{0.67} & 774 \\ 
ADJ & 0.57 & \textbf{0.67} & 0.46 & \textbf{0.58} & 0.51 & \textbf{0.62} & 357 \\ 
NOUN & 0.60 & \textbf{0.70} & 0.49 & \textbf{0.65} & 0.54 & \textbf{0.67} & 863 \\ 
PRON & 0.33 & \textbf{0.34} & 0.18 & \textbf{0.59} & 0.23 & \textbf{0.43} & 131 \\ 
VERB & 0.50 & \textbf{0.76} & \textbf{0.66} & 0.74 & 0.57 & \textbf{0.75} & 361 \\ 
ADP & 0.40 & \textbf{0.41} & 0.46 & \textbf{0.56} & 0.43 & \textbf{0.47} & 325 \\ 
\_ & 0.05 & \textbf{0.08} & 0.01 & \textbf{0.01} & 0.02 & \textbf{0.02} & 98 \\ 
CCONJ & 0.51 & \textbf{0.64} & 0.56 & \textbf{0.64} & 0.53 & \textbf{0.64} & 137 \\ 
NUM & 0.24 & \textbf{0.36} & 0.57 & \textbf{0.63} & 0.34 & \textbf{0.46} & 54 \\ 
PROPN & 0.26 & \textbf{0.34} & 0.70 & \textbf{0.84} & 0.38 & \textbf{0.48} & 114 \\ 
AUX & 0.31 & \textbf{0.60} & 0.12 & \textbf{0.41} & 0.17 & \textbf{0.49} & 190 \\ 
SCONJ & 0.41 & \textbf{0.56} & 0.44 & \textbf{0.70} & 0.42 & \textbf{0.62} & 64 \\ 
X & 0.00 & 0.00 & 0.00 & 0.00 & 0.00 & 0.00 & 64 \\ 
PART & 0.00 & \textbf{0.01} & 0.00 & \textbf{0.03} & 0.00 & \textbf{0.02} & 29 \\ 
\midrule
\textbf{Macro Avg} & 0.36 & \textbf{0.47} & 0.38 & \textbf{0.51} & 0.35 & \textbf{0.47} & 4131 \\ 
\textbf{Weighted Avg} & 0.50 & \textbf{0.62} & 0.47 & \textbf{0.59} & 0.48 & \textbf{0.60} & 4131 \\ 
\bottomrule
\end{tabular}
\caption{POS tagging per entity, comparing \llama and \gpt with 0-shot learning.}
\label{tab:apx:results_pos}
\end{table*}

\begin{table*}[h!]
\centering\small
\begin{tabular}{lccccccc}
\toprule
\textbf{Author} & \multicolumn{2}{c}{\textbf{Precision}} & \multicolumn{2}{c}{\textbf{Recall}} & \multicolumn{2}{c}{\textbf{F1-Score}} & \textbf{Support} \\ 
\cmidrule(lr){2-3} \cmidrule(lr){4-5} \cmidrule(lr){6-7}
& \llama & \gpt & \llama & \gpt & \llama & \gpt & \\ 
\midrule
Thanasis Triaridis & 0.22 & \textbf{0.23} & 0.57 & \textbf{0.61} & 0.32 & \textbf{0.34} & 28 \\ 
Rania Synodinou & 0.14 & \textbf{0.35} & 0.08 & \textbf{0.50} & 0.11 & \textbf{0.41} & 12 \\ 
Kostas Voulazeris & 0.00 & 0.00 & 0.00 & 0.00 & 0.00 & 0.00 & 17 \\ 
Dimitris Tzouvalis & \textbf{1.00} & 0.36 & \textbf{0.31} & 0.31 & \textbf{0.47} & 0.33 & 13 \\ 
Evridiki Amanatidou & \textbf{0.67} & 0.50 & 0.29 & \textbf{0.43} & \textbf{0.40} & 0.46 & 7 \\ 
Frinta Kritsotaki & 0.00 & 0.00 & 0.00 & 0.00 & 0.00 & 0.00 & 5 \\ 
Panos Koliopoulos & 0.00 & \textbf{0.33} & 0.00 & \textbf{0.06} & 0.00 & \textbf{0.10} & 17 \\ 
Yiannis Andamis & 0.00 & \textbf{0.12} & 0.00 & \textbf{0.04} & 0.00 & \textbf{0.06} & 23 \\ 
Giorgos S. Kokkinos & 0.18 & \textbf{0.31} & \textbf{0.75} & 0.62 & 0.29 & \textbf{0.42} & 8 \\ 
Manos Kounougakis & 0.33 & \textbf{1.00} & \textbf{0.20} & 0.20 & 0.25 & \textbf{0.33} & 5 \\ 
Eleni Semertzidou & 0.00 & 0.00 & 0.00 & 0.00 & 0.00 & 0.00 & 5 \\ 
Panos A. Zervas & 0.00 & \textbf{0.17} & 0.00 & \textbf{0.14} & 0.00 & \textbf{0.15} & 7 \\ 
Lakis Fourouklas & 0.00 & 0.00 & 0.00 & 0.00 & 0.00 & 0.00 & 5 \\ 
Vasileios Kappas & 0.00 & 0.00 & 0.00 & 0.00 & 0.00 & 0.00 & 5 \\ 
Paschalis Papavasileiou & 0.80 & \textbf{1.00} & \textbf{0.80} & 0.80 & 0.80 & \textbf{0.89} & 5 \\ 
Plato & 0.75 & \textbf{0.78} & 0.75 & \textbf{0.88} & 0.75 & \textbf{0.82} & 8 \\ 
Aeschylus & \textbf{1.00} & \textbf{1.00} & \textbf{0.40} & 0.20 & \textbf{0.57} & 0.33 & 5 \\ 
\midrule
\textbf{Macro Avg} & 0.30 & \textbf{0.36} & 0.24 & \textbf{0.28} & 0.23 & \textbf{0.27} & 175 \\ 
\textbf{Weighted Avg} & 0.25 & \textbf{0.30} & 0.24 & \textbf{0.29} & 0.20 & \textbf{0.25} & 175 \\ 
\bottomrule
\end{tabular}
\caption{Authorship Attribution evaluation, comparing \llama and \gpt with 0-shot learning.}
\label{tab:apx:results_authorship}
\end{table*}

\end{document}